# Autonomous development and learning in artificial intelligence and robotics: Scaling up deep learning to human–like learning

Pierre-Yves Oudeyer

*Inria and Ensta Paris-Tech, France, 33405 Talence, France*
[pierre-yves.oudeyer@inria.fr](pierre-yves.oudeyer@inria.fr)
[http://www.pyoudeyer.com](http://www.pyoudeyer.com)

***Abstract:*** *Autonomous lifelong development and learning is a fundamental capability of humans, differentiating them from current deep learning systems. However, other branches of artificial intelligence have designed crucial ingredients towards autonomous learning: curiosity and intrinsic motivation, social learning and natural interaction with peers, and embodiment. These mechanisms guide exploration and autonomous choice of goals, and integrating them with deep learning opens stimulating perspectives.*

Deep learning (DL) approaches made great advances in artificial intelligence, but are still far away from human learning. As argued convincingly by Lake et al., differences include human capabilities to learn causal models of the world from very little data, leveraging compositional representations and priors like intuitive physics and psychology. However, there are other fundamental differences between current DL systems and human learning, as well as technical ingredients to fill this gap, that are either superficially, or not adequately, discussed by Lake et al.

These fundamental mechanisms relate to *autonomous development and learning*. They are bound to play a central role in artificial intelligence in the future. Current DL systems require engineers to manually specify a task-specific objective function for every new task, and learn through off-line processing of large training databases. On the contrary, humans learn autonomously open-ended repertoires of skills, deciding for themselves which goals to pursue or value, and which skills to explore, driven by intrinsic motivation/curiosity and social learning through natural interaction with peers. Such learning processes are incremental, online, and progressive. Human child development involves a progressive increase of complexity in a curriculum of learning where skills are explored, acquired, and built on each other, through particular ordering and timing. Finally, human learning happens in the physical world, and through bodily and physical experimentation, under severe constraints on energy, time, and computational resources.

In the two last decades, the field of Developmental and Cognitive Robotics (Cangelosi and Schlesinger, 2015; Asada et al., 2009), in strong interaction with developmental psychology and neuroscience, has achieved significant advances in computational modeling of mechanisms of autonomous development and learning in human infants, and applied them to solve difficult AI problems. These mechanisms include the interaction between several systems that guide active exploration in large and open environments: curiosity, intrinsically motivated reinforcement learning (Schmidhuber, 1991; Oudeyer et al., 2007; Barto, 2013), and goal exploration (Baranes and Oudeyer, 2013), social learning and natural interaction (Chernova and Thomaz, 2014; Vollmer et al., 2014), maturation (Oudeyer et al., 2013) and embodiment (Pfeifer et al., 2007). These mechanisms crucially complement processes of incremental online model building (Nguyen and Peters, 2011), as well as inference and representation learning approaches discussed in the target article.

***Intrinsic motivation, curiosity and free play***. For example, models of how motivational systems allow children to choose which goals to pursue, or which objects or skills to practice in contexts of free play, and how this can impact the formation of developmental structures in lifelong learning, have flourished in the last decade (Baldassarre and Mirolli, 2013; Gottlieb et al., 2013). In depth models of intrinsically motivated exploration, and their links with curiosity, information-seeking and the "child-as-a-scientist" hypothesis (see Gottlieb et al., 2013 for a review), have generated new formal frameworks and hypotheses to understand their structure and function. For example, it was shown that intrinsically motivated exploration, driven by maximization of learning progress (i.e. maximal improvement of predictive or control models of the world, see Schmidhuber, 1991; Oudeyer et al., 2007;) can self-organize long-term developmental structures, where skills are acquired in an order and with timing that shares fundamental properties with human development (Oudeyer and Smith, 2016). For example, the structure of early infant vocal development self-organizes spontaneously from such intrinsically motivated exploration, in interaction with the physical properties of the vocal systems (Moulin-Frier et al., 2014). New experimental paradigms in psychology and neuroscience were recently developed and support these hypotheses (Kidd, 2012; Baranes et al., 2014).

These algorithms of intrinsic motivation are also highly efficient for multitask learning in high-dimensional spaces. In robotics, they allow efficient stochastic selection of parameterized experiments and tasks, enabling to collect data and update skill models incrementally, through automatic and online generation of a learning curriculum. Such active control of the growth of complexity, enables robots with high-dimensional continuous action spaces to learn omnidirectional locomotion on slippery surfaces, and versatile manipulation of soft objects (Baranes and Oudeyer, 2013), or hierarchical control of objects through tool use (Forestier and Oudeyer, 2016). Recent work in deep reinforcement learning has included some of these mechanisms to solve difficult reinforcement learning problems, with rare or deceptive rewards (Bellemare et al., 2016; Kulkarni et al., 2016), as learning multiple (auxiliary) tasks in addition to the target task simplifies the problem (Jaderberg et al., 2016). However, there are many unstudied synergies between models of intrinsic motivation in developmental robotics and Deep RL systems, e.g. curiosity-driven selection of parameterized problems (Baranes and Oudeyer, 2013), of learning strategies (Lopes and Oudeyer, 2012), and combinations between intrinsic motivation and social learning (Nguyen and Oudeyer, 2013), have not yet been integrated with deep learning.

*Embodied self-organization.* The key role of physical embodiment in human learning has also been extensively studied in robotics, and yet it is out of the picture in current deep learning research. The physics of bodies and their interaction with their environment, can spontaneously generate structure guiding learning and exploration (Pfeifer and Bongard, 2007). For example, mechanical legs reproducing essential properties of human leg morphology, generate human-like gaits on mild slopes without any computation (Collins et al., 2005), showing the guiding role of morphology in infant learning of locomotion (Oudeyer, 2016). Yamada et al. (2010) developed a series of models showing that hand-face touch behaviours in the fœtus, and hand looking in the infant, self-organize through interaction of a non-uniform physical distribution of proprioceptive sensors across the body with basic neural plasticity loops. Work on low-level muscle synergies also showed how low-level sensorimotor constraints could simplify learning (Flash and Hochner, 2005).

*Human learning as a complex dynamical system.* Deep learning architectures often focus on inference and optimization. Although these are essential, developmental sciences suggested many times that learning happens through complex dynamical interaction among systems of inference, memory, attention, motivation, low-level sensorimotor loops, embodiment, and social interaction. Although some of these ingredients are part of current DL research, (e.g. attention and memory), the integration of other key ingredients of autonomous learning and development opens stimulating perspectives for scaling up to human learning.